\PassOptionsToPackage{table}{xcolor}
\documentclass[sigconf=true, nonacm=true, review=false, anonymous = false,]{acmart}

\usepackage{booktabs} 

\usepackage{amsmath}
\usepackage{xspace}
\usepackage{xcolor}
\usepackage{dsfont}
\usepackage{rotating}
\usepackage{makecell}
\usepackage{multirow}

\usepackage{amsthm}
\usepackage{amsmath}
\renewcommand{\epsilon}{\varepsilon}



\begin{document}
\title[Modular Differential Evolution]{Modular Differential Evolution}

\author{Diederick Vermetten}
\affiliation{
  \institution{Leiden Institute for Advanced Computer Science}
  \city{Leiden}
  \country{The Netherlands}
}
\email{d.l.vermetten@liacs.leidenuniv.nl}

\author{Fabio Caraffini}
\affiliation{
  \institution{Swansea University}
  \city{Swansea}
  \country{United Kingdom}
}
\email{fabio.caraffini@swansea.ac.uk}

\author{Anna V. Kononova}
\affiliation{
  \institution{Leiden Institute for Advanced Computer Science}
  \city{Leiden}
  \country{The Netherlands}
}
\email{a.kononova@liacs.leidenuniv.nl}

\author{Thomas B{\"a}ck}
\affiliation{
  \institution{Leiden Institute for Advanced Computer Science}
  \city{Leiden}
  \country{The Netherlands}
}
\email{t.h.w.back@liacs.leidenuniv.nl}

\renewcommand{\shortauthors}{D. Vermetten et al.}

\begin{abstract}

New contributions in the field of iterative optimisation heuristics are often made in an iterative manner. Novel algorithmic ideas are not proposed in isolation, but usually as an extension of a preexisting algorithm. Although these contributions are often compared to the base algorithm, it is challenging to make fair comparisons between larger sets of algorithm variants. This happens because even small changes in the experimental setup, parameter settings, or implementation details can cause results to become incomparable. 
Modular algorithms offer a way to overcome these challenges. By implementing the algorithmic modifications into a common framework, many algorithm variants can be compared, while ensuring that implementation details match in all versions. 

In this work, we propose a version of a modular framework for the popular Differential Evolution (DE) algorithm. We show that this modular approach not only aids in comparison, but also allows for a much more detailed exploration of the space of possible DE variants. This is illustrated by showing that tuning the settings of modular DE vastly outperforms a set of commonly used DE versions which have been recreated in our framework. We then investigate these tuned algorithms in detail, highlighting the relation between modules and performance on particular problems. 

\end{abstract}

\maketitle

\section{Introduction}

Science is an iterative process. This is easily seen in action in the evolutionary computation community. 
Most of the contributions made to the state-of-the-art are incremental modifications on a set of core algorithms. This process of researchers building upon the work of others is highly beneficial to the community as a whole, allowing further specialisation of algorithms. However, there are some obstacles to the integration and comparison of many of these proposed modifications.

One particular issue is the fact that algorithms can be inherently challenging to implement. Inconsistencies in the description, ignored edge cases, and even potential bugs can have a significant impact on the behaviour of an algorithm and the interpretation of results~\cite{DBLP:conf/emo/Brockhoff15, biedrzycki2021comparison}. Issues such as these have raised questions regarding the reproducibility of research in computer science as a whole, and evolutionary computation is no exception~\cite{lopez2021reproducibility}. Because of this, comparing different variants of algorithms can be difficult to do fairly. Since researchers often implement the underlying algorithm from scratch, to then add their proposed modification (and in most cases a selected set of other algorithm variants for comparison), clear comparisons are often hard to find.

In an ideal setting, the community would maintain standardised implementations of core algorithms and the proposed modifications would be compared against the same set of state-of-the-art algorithm variants. Unfortunately, this might still be an impossible goal. However, algorithm modifications can still be fairly compared, as long as they are implemented in one common framework. This can be achieved through modular algorithms. 
From one common core algorithm, the variants are implemented as modules which can easily be swapped out. 

The ideas behind modular algorithms have been around for decades~\cite{cahon2004paradiseo, lopez2012automatic, lukasiewycz2011opt4j}, but although they have been shown to be extremely useful~\cite{dreo2021paradiseo}, their adoption in evolutionary computation has been relatively slow. In recent years, several new modular implementations of popular algorithms have been released, including the modular CMA-ES~\cite{de2021tuning,van_rijn_evolving_2016} and the Particle Swarm Optimisation framework~\cite{camacho2021pso}. These works highlight the benefits of modular algorithms not only for fair comparisons, but also hint at the potential to study interactions between modules. 

In this work, we propose a first step towards a modular version of Differential Evolution (DE), a heuristic originally introduced in \cite{bib:Storn1995,bib:Price1997} to optimise a single-objective real-valued fitting problem, and whose design took into consideration elements from evolutionary algorithms and swarm intelligence optimisation (see \cite{bib:ISBDE,vermetten2022analysis} for some insights on these aspects) and a simple core mechanism based on computing difference vectors through linear combinations of candidate solutions. DE has been around for almost 30 years and its popularity means that a wide variety of modifications have been proposed over the years~\cite{DAS20161}. However, when comparing the benchmark data, the relative benefits of many of these modifications seem to vary widely. Our objective is to provide an initial analysis of the performance of a set of 14 independent modules. This does not cover the full space of DE variants, but nonetheless highlights the potential of modular algorithms to aid in understanding the contributions made by these algorithmic variations.

\section{Differential Evolution}\label{sec:DE}
\begin{table*}[]
    \centering
    \caption{Available modules and parameters, their type (`c' for categorical, `i' for integer or `r' for real) and their domain. The choices shown in bold correspond to the default settings. For the numerical parameters, the default values are added after their domain. The `Shorthand' column indicates the names used for these modules in the figures throughout this paper.}\label{tab:module-list}
    \begin{tabular}{lllcp{80mm}}
        \textbf{Operation} & \textbf{Module Name} & \textbf{Shorthand} & \textbf{Type} & \textbf{Domain} \\ \hline \hline
        Initialization & Base sampler & \texttt{Sampler} & c & \{`gaussian', `sobol', `halton',\textbf{ `uniform'}\} \\ 
        Initialization & Oppositional initialisation & \texttt{Opposition} & c & \{true,\textbf{ false}\} \\\hline
        Mutation & Base vector & \texttt{Base} & c & \{\textbf{`rand'}, `best', `target'\} \\
        Mutation & Reference vector & \texttt{Ref} & c & \{\textbf{none}, `pbest', `best', `rand'\} \\
        Mutation & Number of differences & \texttt{Diffs} & c & \{\textbf{1}, 2\} \\
        Mutation & Use weighted F & \texttt{WeightedF} & c & \{true,\textbf{ false}\} \\
        Mutation & Use archive & \texttt{Archive} & c & \{true, \textbf{false}\} \\\hline
        Crossover & Crossover method & \texttt{Crossover} &  c & \{\textbf{`bin'}, `exp'\} \\
        Crossover & Eigenvalue transformation & \texttt{EigenX} & c & \{true, \textbf{false}\} \\\hline
        Bound correction & Bound correction & \texttt{SDIS} & c & \{none, \textbf{'saturate'}, `unif-resample', `COTN', `toroidal', `mirror', `hvb', `expc-target', `expc-center', `exps'\} \\
     \hline      
        Adaptation & F adaptation method & \texttt{AdaptF} & c & \{\textbf{none}, `shade', `shade-modified', `jDE'\} \\
        Adaptation & CR adaptation method & \texttt{AdaptCR} & c & \{\textbf{none}, `shade', `jDE'\} \\
        Adaptation & Population size reduction & \texttt{LPSR} & c & \{true, \textbf{false}\} \\ 
        Adaptation & Use JSO caps for F and CR & \texttt{Caps} & c & \{true, \textbf{false}\} \\\hline

        Parameter & Population size & $\lambda$ & i & \{4, \dots, 200\} ($\boldsymbol{{4 + }\lfloor(3 \log(D))\rfloor}$)\\
        Parameter & Scale factor & $F$ & r & [0, 2]  (\textbf{0.5})\\
        Parameter & Crossover rate & $CR$ & r & [0, 1] (\textbf{0.5}) \\ \hline
    \end{tabular}
\end{table*}

The original DE framework offers a somewhat modular structure by design, where DE variants are characterised by the notation \texttt{DE/x/y/z-SDIS}. 
Here, we stress the importance of considering the SDIS (Strategy for Dealing with Infeasible Solutions) as a non-optional operator of a heuristic algorithm \cite{DEOTB}. To understand this notation and describe the algorithm in a compact way, some DE jargon described below is needed. 

DE is a population-based algorithm where one iteration (i.e. a generation) is completed by perturbing candidate solutions in the population one at a time - from the first to the last individual. This process starts from some initial population and is iterated for multiple generations depending on the available computational budget. The order in which individuals are selected to undergo the variation operators is maintained. In this context, a selected solution is referred to as the \texttt{target} individual $\mathbf{x}_t$. This individual undergoes a crossover operator \texttt{z}, which first requires the preparation of a second individual, called the \texttt{mutant} individual $\mathbf{x_m}$, to produce an offspring $\mathbf{x}_o$ called the \texttt{trial} vector. This solution might contain infeasible components inherited from \texttt{mutant} and therefore must be fed to a \texttt{SDIS} operator. Immediately after generating a feasible \texttt{trial} solution, \texttt{target} and \texttt{trial} compete to enter the next population (note that the swap takes place only after the generation cycle is complete in the original DE framework - i.e., after each individual has produced a \texttt{trial} solution). 

DE mutation operator \texttt{x/y} is executed to obtain a \texttt{mutant} solution. This is the operator that gives the name to the algorithm, as it is based on the idea of `moving' a point (usually a randomly chosen solution from the population, i.e. \texttt{x=rand}, or the best so far individual, i.e. \texttt{x=best}, or a combination of individuals that can include the \texttt{target} - see \cite{bib:DEbook} and Section \ref{sec:mut}) by adding a number \texttt{y} of scaled `difference' vectors. This results in a linear combination of individuals because the difference vectors are obtained by subtracting two randomly chosen individuals. Note that all solutions involved in \texttt{x/y} must be distinct and feasible individuals or combinations of individuals. Furthermore, in the case where \texttt{x} is chosen as a vector originating from one individual to another, for example \texttt{rand-to-best} or \texttt{current-to-rand}, some recombination is already present at the mutation level, leaving the option of dropping the crossover operator. To function, this simple algorithmic structure only requires setting a population size $\lambda$, a scale factor $F\in [0,2]$ for the mutation operator, and a crossover rate $CR\in(0,1]$ for the crossover. A detailed pseudocode of this framework can be found in \cite{DEOTB} and for more details and examples of DE operators, we refer to \cite{DAS20161}.

\section{Included Modules}

Similar to other heuristic optimisers, DE naturally lends itself to a reformulation as a modular algorithm made up of a number of connected modules/operators where every independently made choice for a module is fully compatible with all choices for other modules. In fact, previous work has shown the usefulness of considering these operators as independent modules, e.g. to rigorously analyse the impact of the crossover operator~\cite{campelo2016experimental}. In this paper, we use this modularity to create a framework which we call \textit{Modular DE} where a full combinatorial range of modules is available for each algorithm component, see Table~\ref{tab:module-list}.

 \begin{table*}
    \centering
    \caption{Set of 11 commonly used DE variants and the way they are implemented in modular DE. Empty cells indicate default values are used.}\label{tab:common_de}
    \begin{tabular}{lllcccl}
        \textbf{Name/Author} & & \textbf{Mutation Settings} & \textbf{F} & \textbf{CR} & $\boldsymbol{\lambda}$ & \textbf{Other Settings} \\ \hline \hline
        L-SHADE & \multirow{2}{*}{\citep{bib:tanabe2013}} & \texttt{Base} : target, \texttt{Ref} : pbest & \multicolumn{2}{c}{adaptive} & $18\cdot D$ &  \texttt{LPSR} : true, \texttt{Archive} : true, \texttt{AdaptF} : shade, \texttt{AdaptCR} : shade\\
        SHADE &  & \texttt{Base} : target, \texttt{Ref} : pbest & \multicolumn{2}{c}{adaptive} & $10\cdot D$ &  \texttt{Archive} : true, \texttt{AdaptF} : shade, \texttt{AdaptCR} : shade\\ \hline
        DAS1 & \multirow{2}{*}{\cite{das2009differential}}   & & $0.8$ & $0.9$ &  $10\cdot D$ \\
        DAS2 &    & \texttt{Base} : target, \texttt{Ref} : best & $0.8$ & $0.9$ &  $10\cdot D$ \\ \hline
        Qin1 & \multirow{4}{*}{\cite{qin2008differential}} & & $0.9$ & $0.9$ &  $50$ \\
        Qin2 &  & & $0.5$ & $0.3$ &  $50$ \\
        Qin3 & & \texttt{Ref} : best & $0.5$ & $0.3$ &  $50$ \\
        Qin4 & & \texttt{Ref} : best, \texttt{Diffs} : 2 & $0.5$ & $0.3$ &  $50$ \\ \hline
        Gamperle1 & \multirow{2}{*}{\cite{gamperle2002parameter}} & \texttt{Ref} : best, \texttt{Diffs} : 2 & $0.45$ & $0.4$ &  $2\cdot D$ \\
        Gamperle2 &  & \texttt{Ref} : best, \texttt{Diffs} : 2 & $0.6$ & $0.9$ &  $2\cdot D$ \\ \hline
        jDE & \cite{bib:brest2006} & & \multicolumn{2}{c}{adaptive} & $100$ & \texttt{AdaptF} : jDE, \texttt{AdaptCR} : jDE \\ \hline
    \end{tabular}
\end{table*}

\subsection{Initialisation}
To create the initial population, we implemented several sampling strategies (\texttt{Sampler}, see Table \ref{tab:module-list}). The most common is to create a uniform distribution across the entire domain. Alternatives are to use other distributions or low-discrepancy sampling methods. We choose to include the Halton and Sobol sequences to represent low-discrepancy sampling and a Gaussian distribution (centred around the origin, with $\sigma=(U-L)/6$, where $U$ and $L$ are the upper and lower bounds, respectively) to represent other kinds of distribution. Furthermore, a previous study has proposed using an oppositional initialisation strategy~\cite{bib:Rahnamayan2008} (\texttt{Opposition}), where each time we generate an individual for the initial population, we also generate its mirror image around the origin. 

\subsection{Mutation}\label{sec:mut}
The mutation operator has been the focus of many modifications of DE, see, e.g. \cite{zhang2009jade,5678831,6046144, DAS20161, 8695569}. To capture the most established mutation variations of the kind \texttt{x/y} (as described in Section \ref{sec:DE}), and to give flexibility in adding new variants, we implement the mutation operator through the combination of 3 modules. The first two modules, namely \texttt{Base} and \texttt{Ref}, help define the strategy \texttt{x}. Note that the reference solution \texttt{Ref} can be set to \texttt{none}, while the \texttt{Base} solution is not optional. In this scenario \texttt{x = Base}. Conversely, when \texttt{Ref} is one of the admissible reference solutions displayed in Table \ref{tab:module-list}, a scaled version of the vector directed from \texttt{target} to the reference point is generated and added to \texttt{Base}, i.e. \texttt{Base + F(Ref-target)}. Therefore, when \texttt{Ref} is not \texttt{none}, one obtains any of the classic strategies of the kind \texttt{x = target-Refs}, plus new ones by varying the base vector.  The third module, namely \texttt{Diffs}, is used to set the number \texttt{y} of difference vectors. 

In addition to this restructuring of the definition of the mutation operator, we implement the option of using \texttt{WeightedF}, which reduces $F$ at the beginning of the search and then increases it towards the end~\cite{brest2017single}.

One more modification makes use of an archive of external solutions, as done, e.g., in \cite{zhang2009jade}, where one of the solutions in the archive is chosen to be part of one of the difference vectors - a scheme that has been shown to lead to improvements in the past and is activated via the module \texttt{Archive}.

\subsection{Crossover}
The classical studies in DE generally consider two types of crossover: binomial (\texttt{z=bin}) and exponential (\texttt{z=exp}) \cite{bib:DEbook}, where the names refer to the distributions used for the probability of exchanging design variables between \texttt{target} and \texttt{mutant}. Both these types of crossover are included in this work.

Furthermore, we also include the option of performing the procedure from \cite{bib:Guo2015}, by activating the eigenvalues transformation module \texttt{EigenX}, which allows using the \texttt{bin} or the \texttt{exp} operator and still maintaining rotational invariant behaviour. This is obtained by producing a covariance matrix from the individuals that make up the current population and diagonalising it with the Jacobi method \cite{bib:Demmel1992} to calculate the eigenvalues and eigenvectors. These are real-valued and form an orthogonal basis (since the covariance matrix is symmetric and surely diagonalisable) and are arranged in a matrix R used to rotate \texttt{target} and \texttt{mutant} before performing the crossover. Note that the obtained \texttt{trial} has to be transformed back to the original coordinate system. This is an easy task, as the conjugate matrix R$^*$ is equivalent to $\text{R}^\text{T}$ in this scenario. Therefore, the multiplication between the transposed transformation matrix $\text{R}^\text{T}$ and the newly generated point returns the desired \texttt{trial}.

\subsection{Boundary Correction}

There exist several mechanisms for boundary correction in the literature that allow us to deal with infeasible solutions. The most used within the DE community can be found in \cite{bib:BIEDRZYCKI2019, kononova2022importance}. For the proposed modular DE framework, we selected a varied set of 10 strategies for box-constrained problems (such as all problems in the BBOB test suite) which are fully described and analysed in such articles, and we refer to~\cite{reproducibiliyt} for implementation details.

\subsection{Parameter Adaptation}

Most state-of-the-art DE variants make use of adaptive parameters. So, in the proposed modular framework we implement adaptation methods for the DE core parameters, namely $F$, $CR$, and $\lambda$. The simplest is \texttt{LPSR}, which linearly reduces the population size over time~\cite{bib:Brest2008b}. For $F$ and $CR$, we implement the adaptation mechanisms of SHADE and jDE \cite{bib:brest2006, bib:tanabe2013}. For $F$, we add an additional mechanism which uses the mean of the memory, instead of generating a different distribution for each individual, in the SHADE's adaptation strategy.

One final option to change the adaptation process is to use JSO caps for $F$ and $CR$ (\texttt{Caps}), which, once activated, caps the values of these two parameters with different thresholds depending on conditions on the used computational budget~\cite{brest2017single}. 

\section{Experimental Setup}
\textbf{Experiment 1} In order to analyse the potential of modular implementation of DE, we recreate a set of 11 \textit{known} versions of DE within our framework (referred to as \textit{common} variants). These algorithms are shown in Table~\ref{tab:common_de}, where all non-default parameters are mentioned. In addition to this, we can create a set of $30$ \textit{single-module variations}: DE versions where all modules are set to their default value, except for one. As such, each non-default module option is enabled in exactly one \textit{single-module} variant. For these \textit{single-module} variants, we set $F=CR=0.7$, and $\lambda=10\cdot D$, based on the recommendations of~\cite{tuningLampinen}.

To benchmark this portfolio, we use the BBOB suite from the COCO platform~\cite{hansen2020coco}. This suite contains 24 noiseless, single-objec\-tive, continuous optimisation problems, each of which can be instantiated with different transformations. These instances aim to provide slight deviations from the original function while preserving the properties of the global landscape. For each DE variant, we collect performance data on all 24 BBOB problems, using \texttt{IOHexperimenter}~\cite{iohexp} for data collection. We perform $50$ runs per function, spread over $10$ instances (5 independent runs per instance). We repeat this for dimensions $D\in\{5,10,20\}$, where we give each run a budget of $50\,000$ function evaluations. 

To evaluate the performance of each algorithm, we opt to use the Empirical Cumulative Distribution Function (ECDF). In particular, we use the \textit{Area Over the ECDF Curve} (AOC) as an \textit{anytime performance} measure~\cite{hansen2022anytime}. Note that we make use of AOC instead of the more common Area Under the Curve (AUC) to keep the interpretation of minimizing the performance metric. To calculate the ECDF and the corresponding AOC, we use a set of $81$ targets logarithmically scaled between $10^8$ and $10^{-8}$. These targets are based on precision (difference between best-so-far $f(x)$ found and the f-value of the global optimum) to allow aggregation across instances. As a last step, we normalise the AOC to lie in $[0,1]$ by dividing the computed value by the total budget. For \textit{interpreting the AOC values}, we should keep in mind that when all targets are hit immediately, we get a value of $0$, which is optimal, while $1$ indicates that none of the selected targets have been hit throughout the optimisation run. 

\begin{figure}[!t]
    \centering
    \includegraphics[width=0.49\textwidth,trim=9mm 10mm 4mm 5mm,clip]{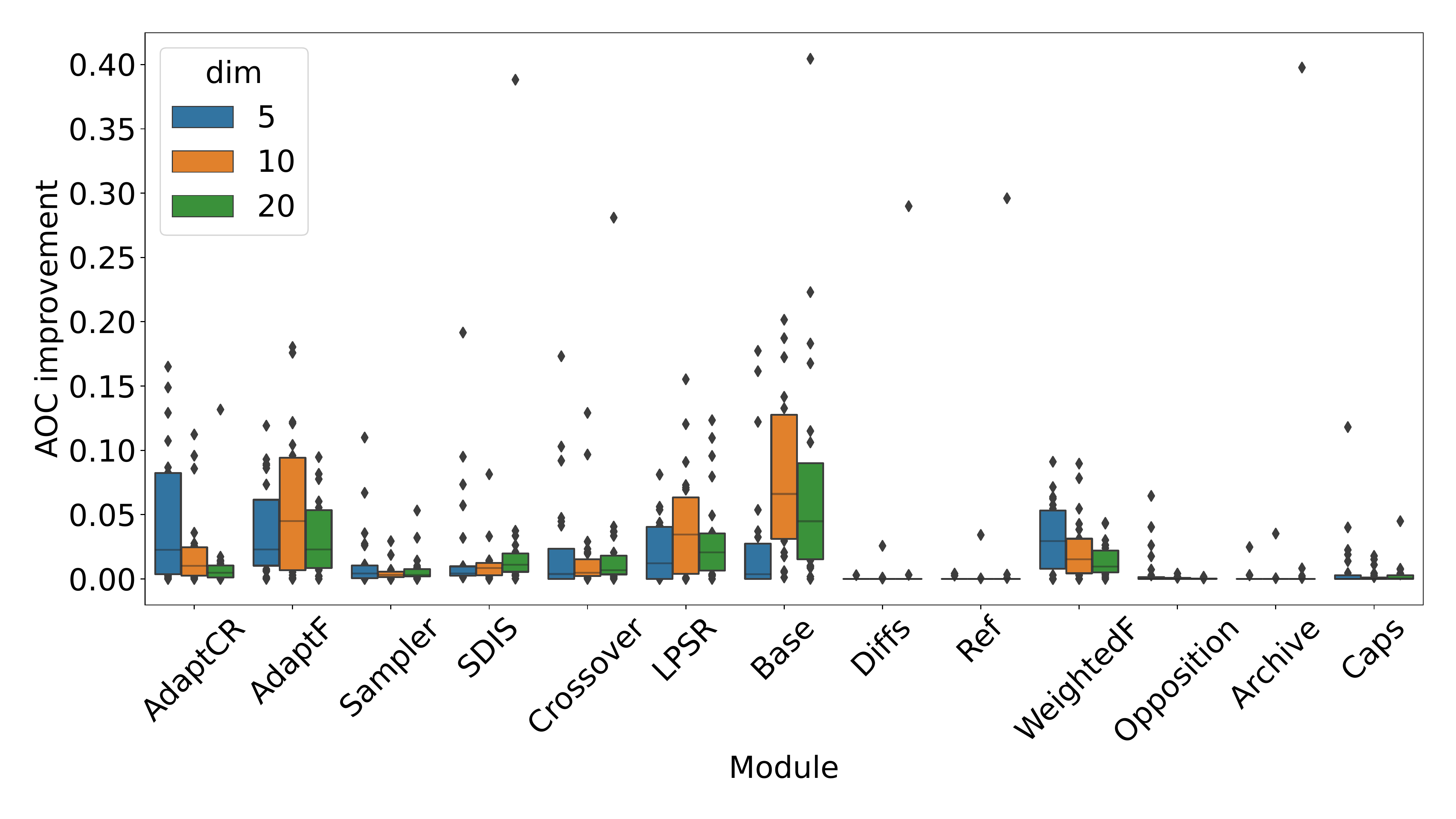}
    \caption{Improvement in AOC over the default setting when selecting the best-performing option for each module.}
    \label{fig:single_module_impr}
\end{figure}

\textbf{Experiment 2} For our second set of experiments, we use the algorithm configuration tool \texttt{irace}~\cite{irace} to tune the performance of the modular DE on the same set of BBOB problems. Each irace run uses a budget of $10\,000$ evaluations, where each evaluation corresponds to running a DE variant with the selected parameter setting. We use the f-test version of irace, with a first-test value of 5, with the other parameters set to their default values.

 We perform 10 independent runs of irace on each function from the BBOB suite, for dimensions $D\in\{5,10,20\}$, where irace has access to the first 5 instances of the function. We set the targets for ECDF to 81 logarithmically spaced values between $10^8$ and $10^{-8}$. 
 We use AOC as the target since it has been shown that the increased signal it captures relative to measures such as Expected Running Time (ERT) can lead to overall performance improvements, even when evaluating the result with a different measure~\cite{ye2022automated}. 
 In addition to these per-function tuning runs, we also perform 10 tuning runs where we tune for aggregated performance over all the functions by setting the irace instance set to the 24 BBOB problems. 
 
 The resulting \textit{elite configurations} for the across-function tuning are validated using the same settings as the DE variants from the first experiment: 5 independent runs on 10 instances of each BBOB problem. For the per-function tuning, we instead perform 5 independent runs on 50 instances of the problem on which the tuning was performed.

\textbf{Reproducibility:} To ensure the reproducibility of our results, the complete set of scripts used for these experiments have been uploaded to a Zenodo repository~\cite{reproducibiliyt}. This repository also contains the resulting irace logs, table with elite configurations, and verification runs in IOHanalyzer format. The notebooks used for analysis and visualisation are also provided. A set of additional figures that could not be included in this paper has been added to Figshare~\cite{reproducibiliyt}.  

\begin{figure}[!t]
    \centering
    \includegraphics[width=0.49\textwidth,trim=9mm 10mm 4mm 5mm,clip]{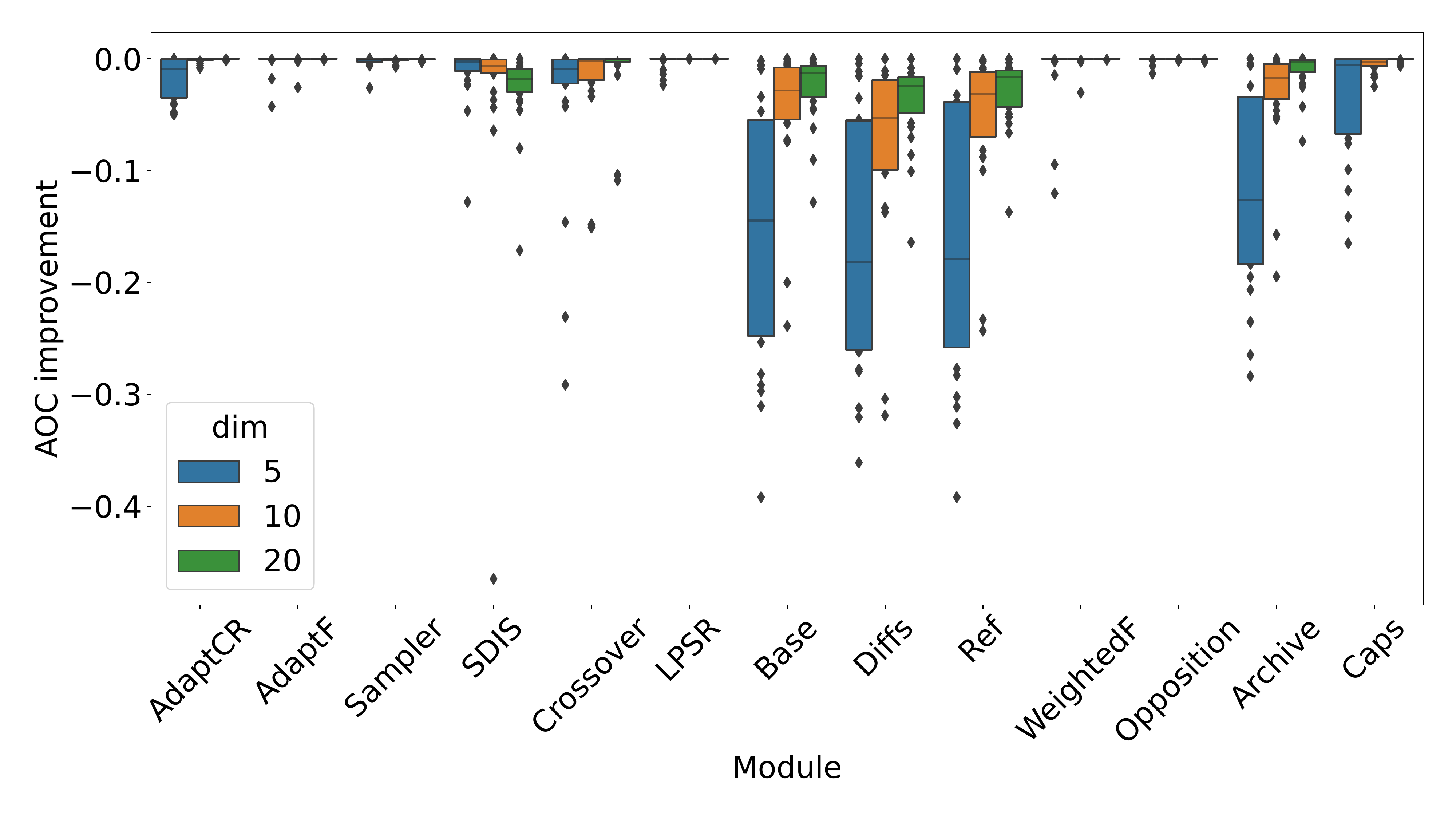}
    \caption{Reduction in AOC over the default setting when selecting the worst-performing option for each module.} 
    \label{fig:single_module_red}
\end{figure}

\begin{figure}[!t]
    \centering
    \includegraphics[width=0.49\textwidth,trim=9mm 11mm 25mm 12mm,clip]{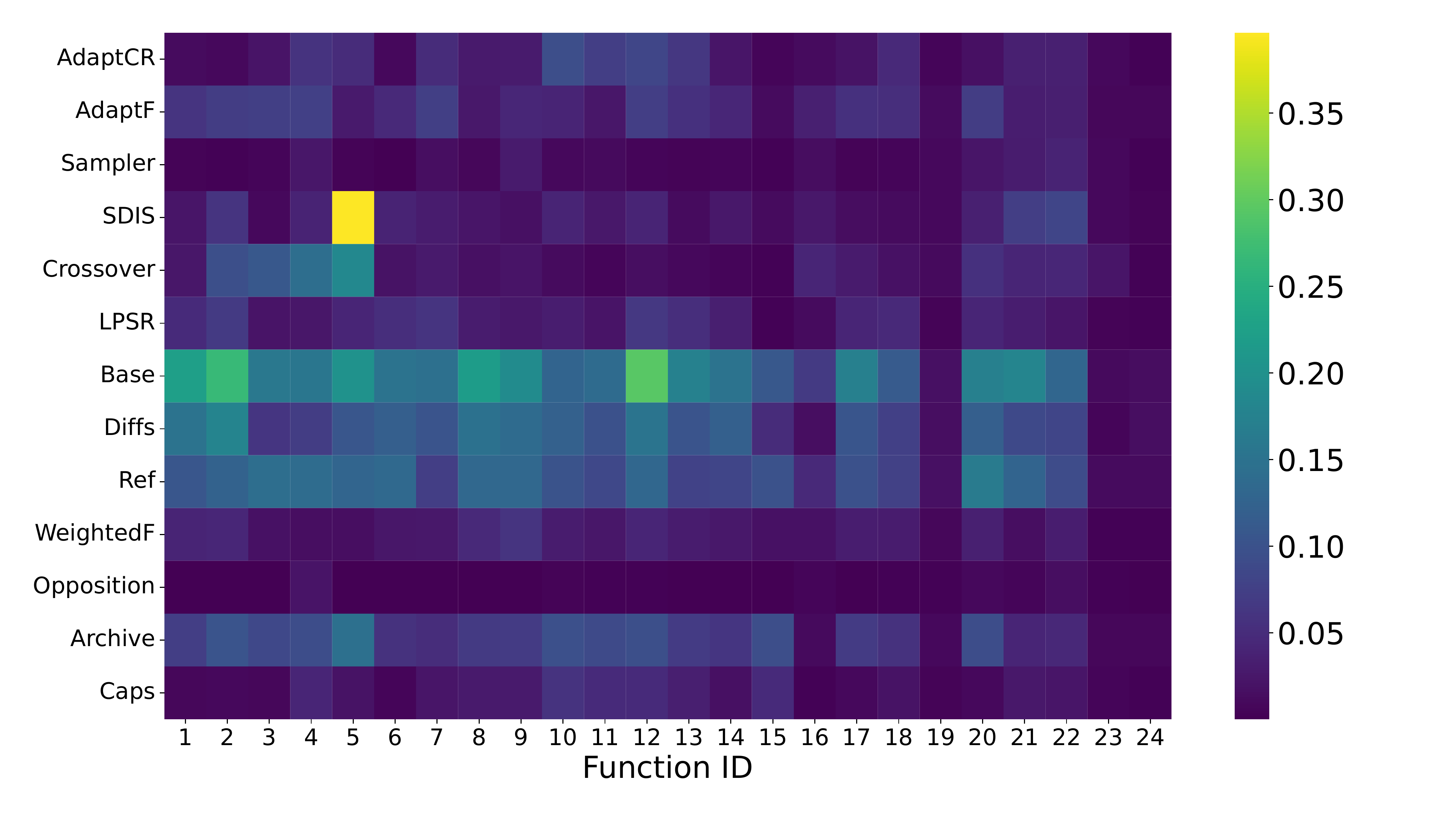}
    \caption{Importance of each module to the AOC on each of the 24 BBOB functions, aggregated over the used dimensions. Importance is calculated as the sum of absolute values from Fi\-gures~\ref{fig:single_module_impr} and \ref{fig:single_module_red}.}
    \label{fig:single_module_importance_heatmap}
\end{figure}

\begin{figure}
    \centering
    \includegraphics[width=0.49\textwidth,trim=9mm 11mm 4mm 5mm,clip]{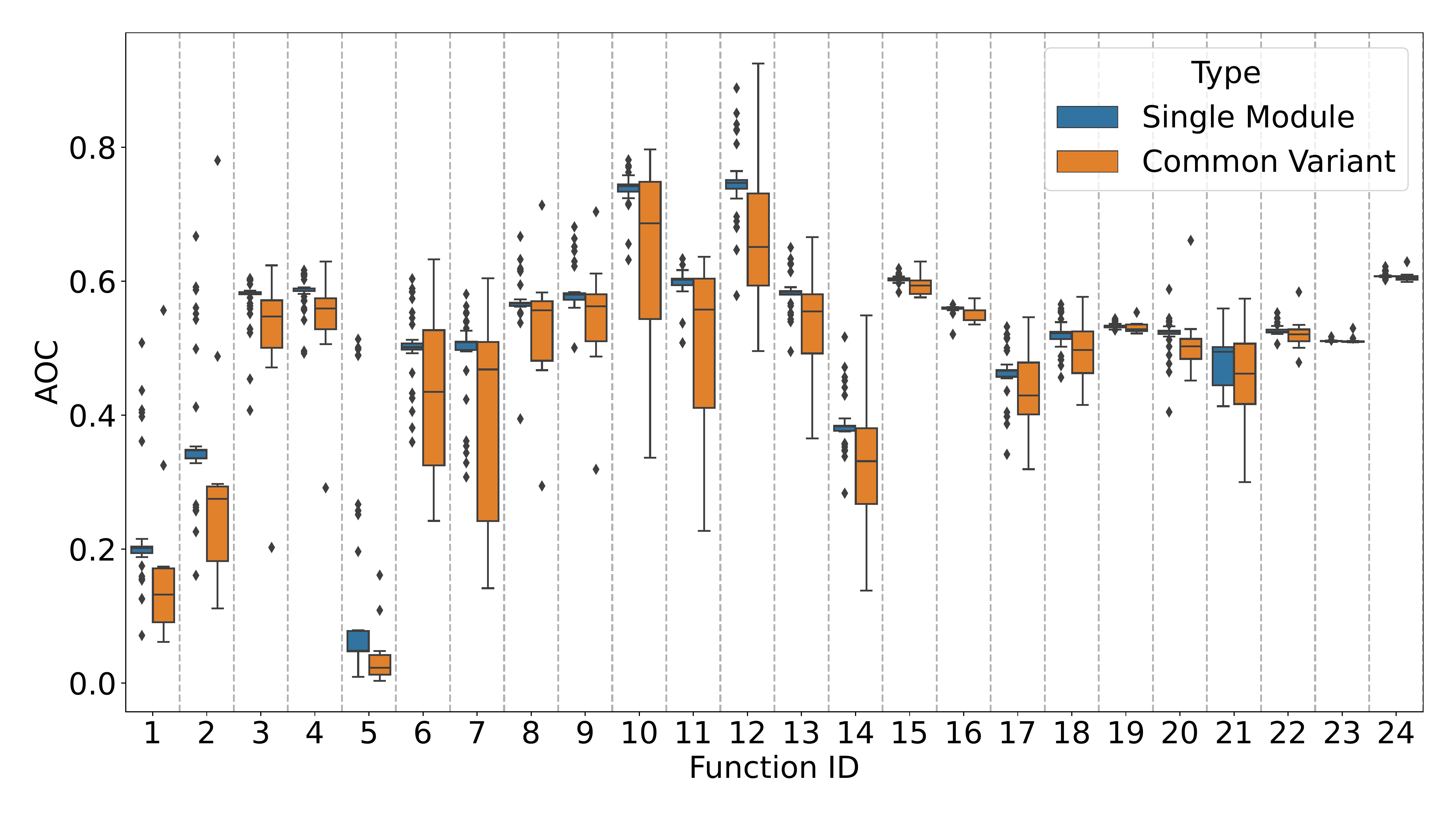}
    \caption{Performance distribution (AOC) of the 30 \textit{single-module} DE variants and the 11 \textit{common} DE variants from Table~\ref{tab:common_de}, for the 10-dimensional BBOB problems.}
    \label{fig:perf_common_vs_sm}
\end{figure}

\section{Single-Module and Common DE Variants}

First, we investigate the \textit{single-module} DE variants, which can be used to illustrate the impact of each module in isolation. We achieve this by comparing the performance of the default DE (all modules at their default value as seen in Table~\ref{tab:module-list}) to the variant with the identified best options enabled for each module. The resulting distribution of improvements is shown in Figure~\ref{fig:single_module_impr}. 

From Figure~\ref{fig:single_module_impr}, we can see that some modules have relatively minor impact when the optimal option is selected independently from any other modules. This is the case for e.g. the number of difference components (\texttt{Diffs}) and the use of an archive population (\texttt{Archive}). In fact, if we instead consider the performance deterioration when making the worst choice for each module, these ones show a significant change over the default setting, as can be seen in Figure~\ref{fig:single_module_red}. The combination of these two figures gives an overall importance of each module, in the sense that if only one module can be modified, some modules will likely have a much larger impact on the overall performance of the algorithm than others. The aggregation of maximum improvements and deteriorations for the selection of different module options is visualized in Figure~\ref{fig:single_module_importance_heatmap}. This figure shows the way in which these module importances are distributed across functions. For some functions, all \textit{single-module} configurations perform similarly poorly, e.g. for F24, so no differences are detected. For most others, differences are present, with a clear impact on the choice of the base vector used for mutation (\texttt{Base}). In general, the mutation modules have relatively more impact than most others. Somewhat surprisingly, the impact of the adaptation methods for F, CR and population size is rather small. This might indicate that these settings work best when combined with other modules or more specific parameter settings. Also worth noting is that boundary correction is usually not impactful, with the exception of F5 (linear slope). For this function, the optimum lies directly on the boundary, so the boundary correction will be triggered often when close to the optimum, and thus have a large impact on the algorithm's performance~\cite{kononova2022importance}. All other BBOB functions are known not to have optima in the relative vicinity of domain boundaries~\cite{long2022bbob}.

\begin{figure}
    \centering
    \includegraphics[width=0.49\textwidth,trim=9mm 11mm 4mm 5mm,clip]{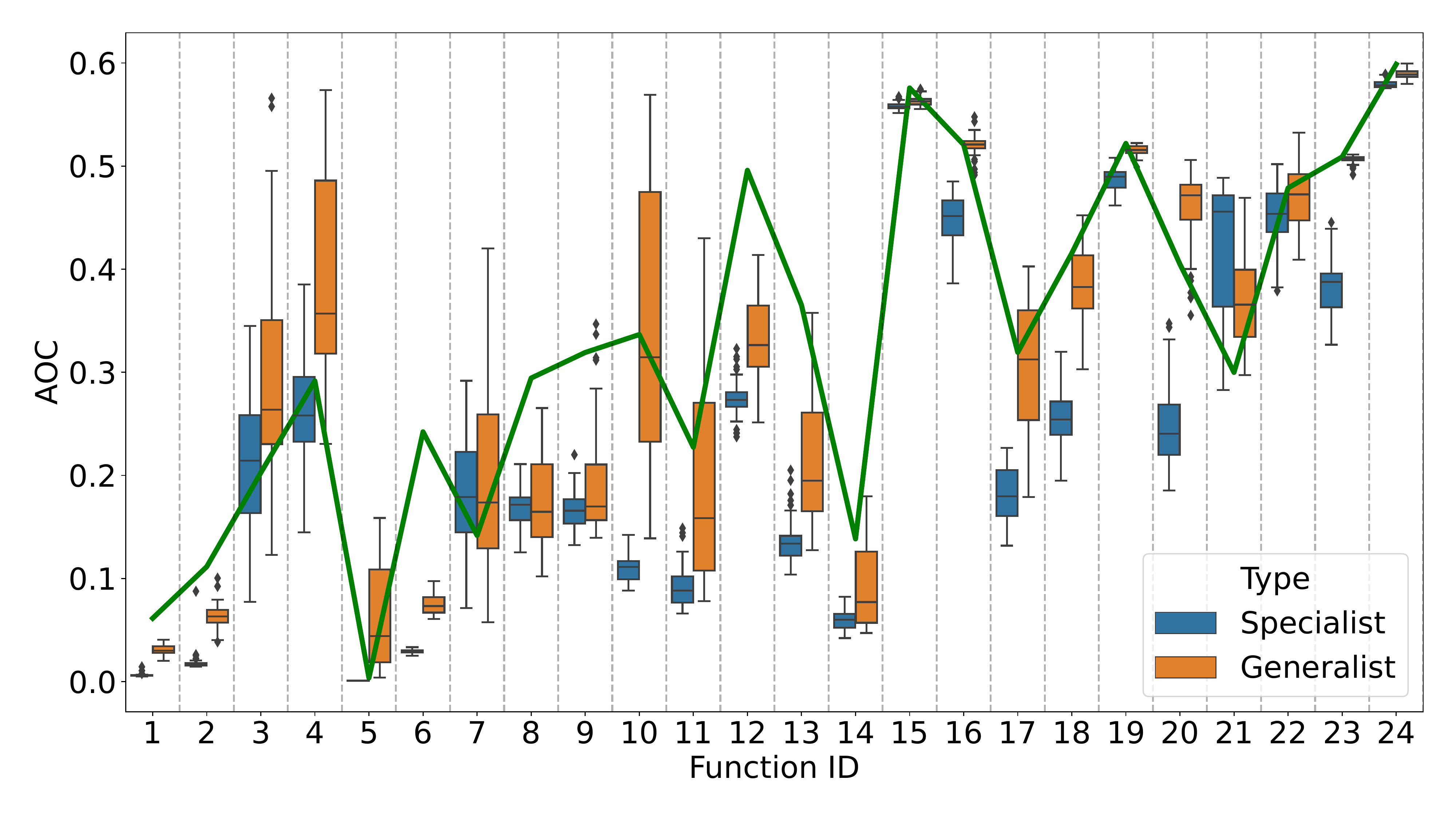}
    \caption{Performance distribution (AOC) of the configurations tuned for an individual function (\textit{specialist}) and the configurations tuned for the full BBOB suite (\textit{generalist}), for the 10-dimensional BBOB problems. The green line shows the best DE version from the union of \textit{single-module} DE and \textit{common} DE variants from Table~\ref{tab:common_de}. }
    \label{fig:spec_vs_gen}
\end{figure}

\begin{figure*}
    \centering
    \includegraphics[width=0.98\textwidth,trim=9mm 11mm 4mm 5mm,clip]{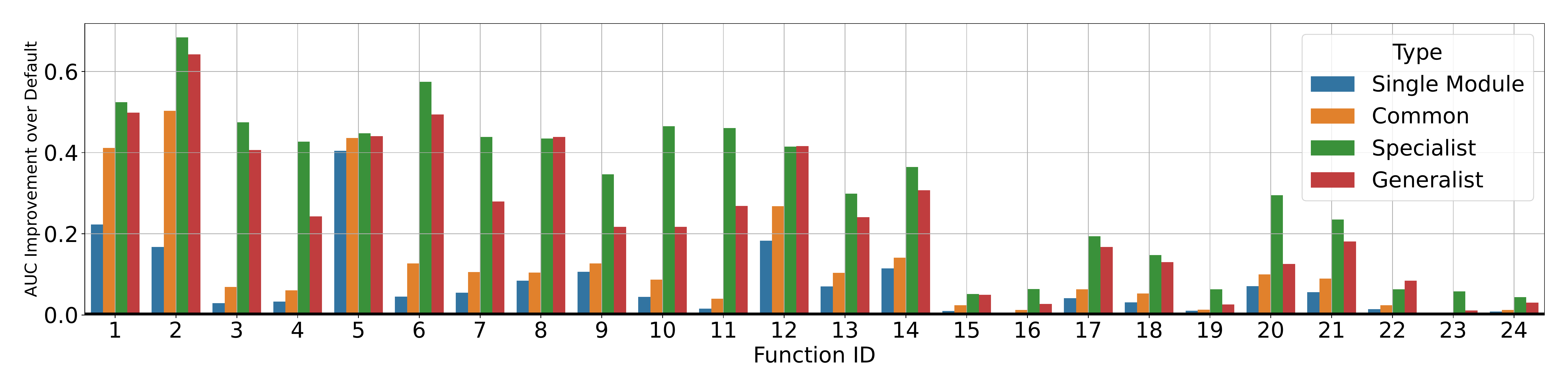}
    \caption{Relative improvement in AOC value between the best configuration of each type and the default setting, in 20D.}
    \label{fig:impact_tuning_20d}
\end{figure*}

To get insight into how hand-crafted DE versions, such as L-SHADE, compare to the \textit{single-module} ones, we look at the performance distributions on the 24 BBOB problems. This is visualised in Figure~\ref{fig:perf_common_vs_sm}. From this figure, we see that there is a fairly wide distribution of performance in both groups. Overall, the \textit{common} DE variants seem to contain better configurations, although the set of configurations is relatively much smaller.

\section{Performance of Tuned DE}
Next, we compare the hand-crafted and \textit{single-module} DE versions to those resulting from tuning the modular DE using irace. The resulting performance on the 10D BBOB problems is visualized in Figure~\ref{fig:spec_vs_gen}. From this figure, we can see that generally, both of the tuned DE settings outperform the hand-crafted ones. As expected, tuning for a particular function improves the performance on that function rather significantly. 

Next, we aim to understand the impact of tuning relative to picking the best configuration from the set of \textit{common} variants. To investigate this, we look at the relative gain in AOC over the default, for each set of configurations (\textit{common} variants, \textit{single-module} variants, \textit{specialists} and \textit{generalists}). For each type, we look at the performance of the best configuration of that type on each function and take the improvement it makes over the default setting. These improvements, for the 20D BBOB functions, are visualized in Figure~\ref{fig:impact_tuning_20d}. From this figure, we can see that the default setting performs particularly poorly on most of the unimodal problems, as even the best \textit{single-module} configuration can outperform it significantly. However, this also shows the additional benefit which can be gained from tuning, which is particularly noticeable e.g. F3 and F4. We should also note that the performance gains shown here are slightly larger than those seen in Figure~\ref{fig:spec_vs_gen}, which in turn are slightly larger than those achieved on the 5D version of these problems. The figures for these other settings can be found on our Figshare repository~\cite{reproducibiliyt}. 

One more important note from Figure~\ref{fig:spec_vs_gen} is the wide distribution of AOC values. For the \textit{generalist} configurations, this is natural, as configurations with different strengths can achieve similar performance when aggregated over the whole BBOB suite, resulting in a large per-function variance when grouped together. However, for the configurations tuned on a single function, the variance on some functions is still clearly visible. This might be caused by the inherent stochasticity of DE which potentially misleads the algorithm configurator when limited samples are available~\cite{vermetten2022analyzing}. 

This might also explain why for F21 one of the hand-crafted DE variants outperforms almost all configurations which were tuned on that function. When considering Figure~\ref{fig:perf_common_vs_sm}, we see that the performance might be considered an outlier, which performs much better than the remaining \textit{common} variants. This observation might indicate that using the \textit{common} DE variants to initialize irace might be able to provide some additional benefits over the current random sampling.

\begin{figure}[!b]
    \centering
    \includegraphics[width=0.49\textwidth,trim=11mm 11mm 20mm 9mm,clip]{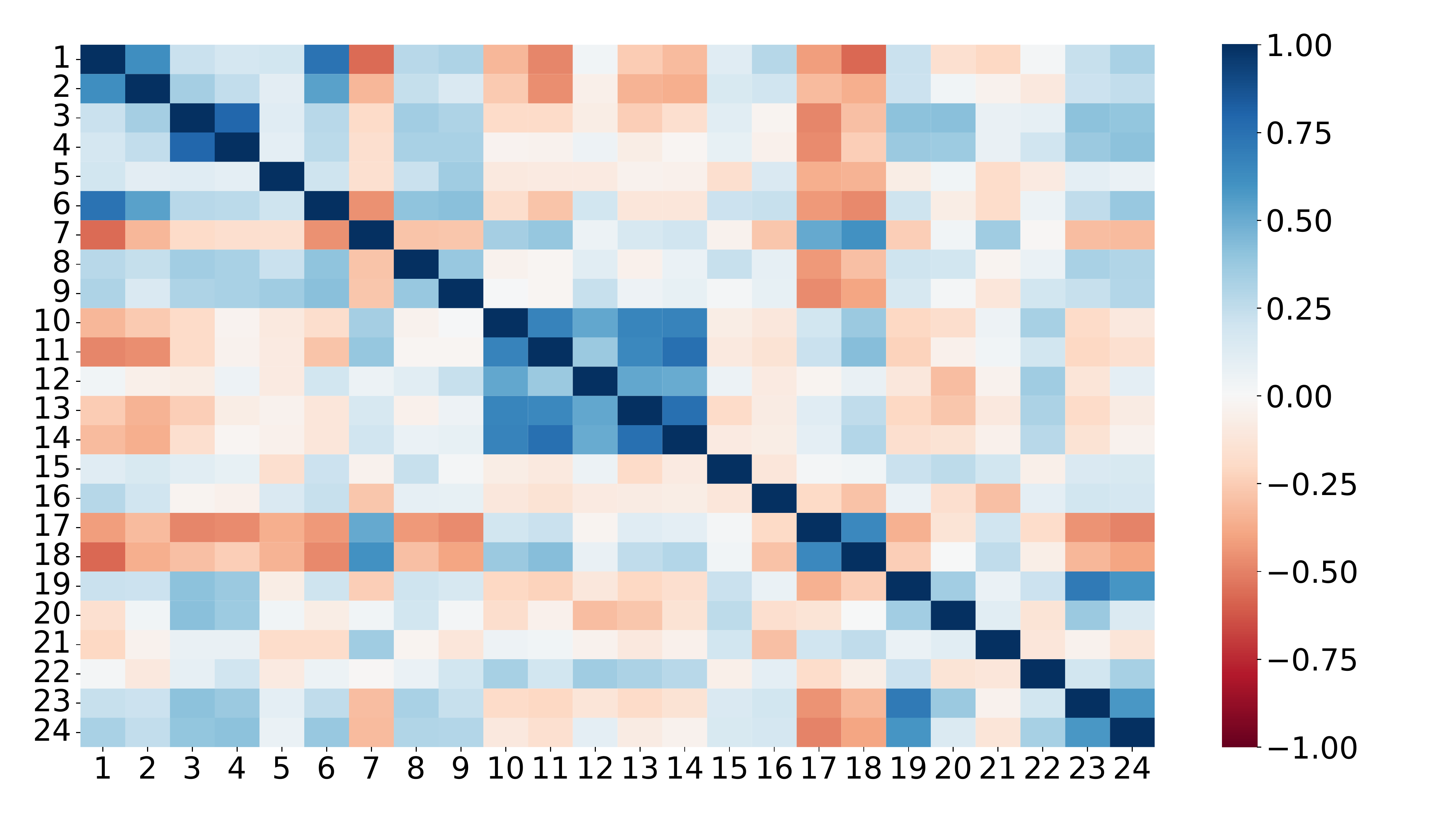}
    \caption{Kendall Tau correlation between the 24 functions of the BBOB suite, measured based on the performance of the set of elites which were tuned on the full BBOB suite in 10D.}
    \label{fig:across_fct_corr}
\end{figure}

Since the variance in performance between \textit{generalist} configurations is large, it would be worthwhile to investigate the correlation between functions, based on the performance of the elite configurations. This is visualised in Figure~\ref{fig:across_fct_corr}. From this figure, we see that several groups of problems seem to appear. This grouping might indicate that several of these functions could be removed from the tuning set. Since the algorithm configuration process works on the basis of ranking, testing on multiple instances where the ranking of configurations is almost equivalent might not be very beneficial. Removing some functions from the training set would allow for a better representation of the function space to be used in each individual race, resulting in potentially improved \textit{generalist} configurations. 

\section{Analysis of Elite Configurations}
Since multiple repetitions of irace are performed for each problem, we have a set of between $10$ and $50$ elite configurations for each setting. By analysing the commonalities between these elites, we can get an overview of the benefit of different parameter settings. This can be done on a global level by aggregating the activations of certain module options across runs and dimensions. 

\begin{figure}[!b]
    \centering
    \includegraphics[width=0.49\textwidth,trim=9mm 11mm 4mm 5mm,clip]{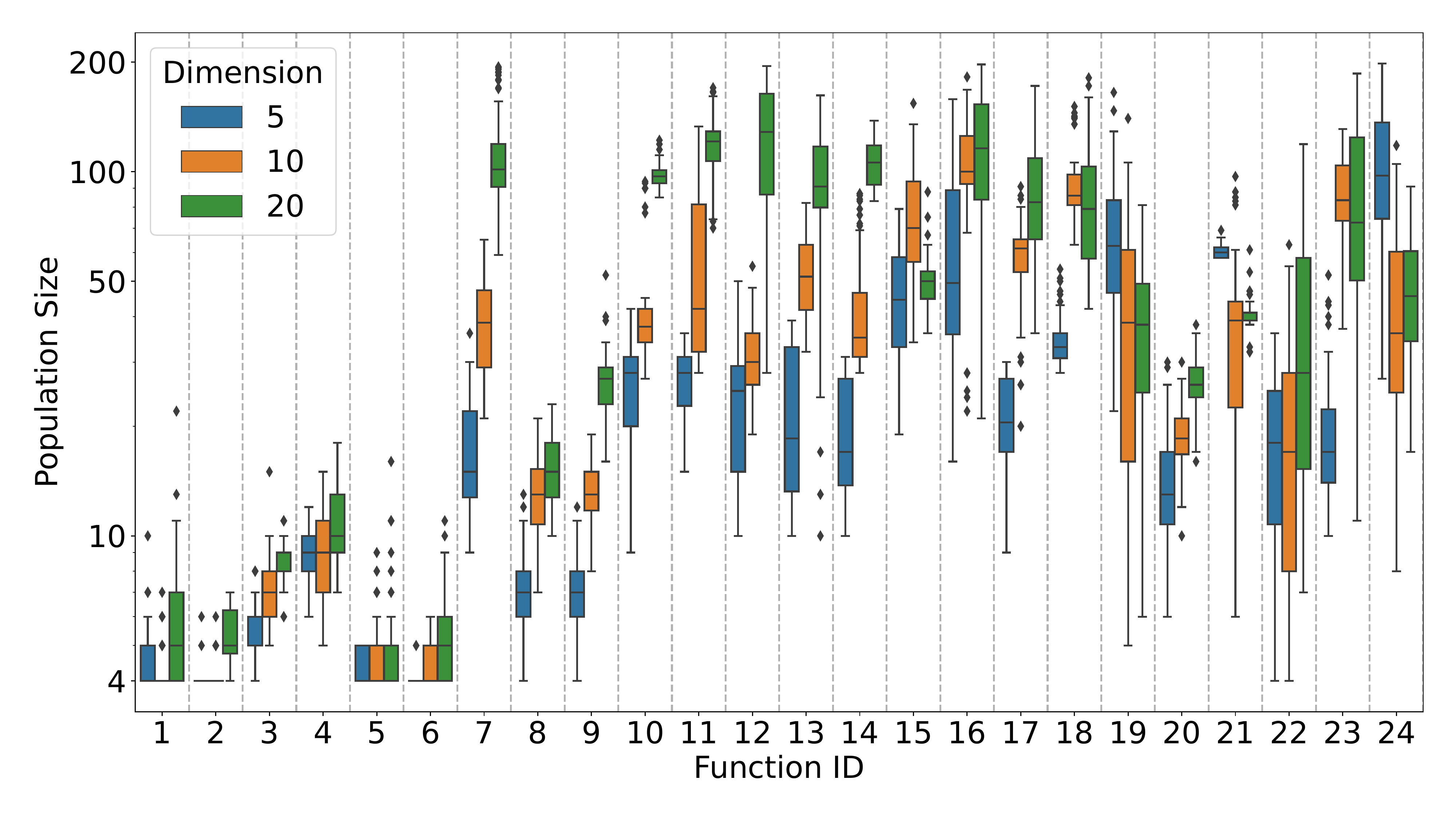}
    \caption{Distribution of population size per dimension in elite configurations (\textit{specialist}).}
    \label{fig:popsize}
\end{figure}

In Figure~\ref{fig:popsize}, we show how the population size changes with respect to functions and dimensionality. In particular, we observe that for the more simple problems, such as the sphere (F1) and ellipsoid (F2), lower population sizes are preferred. In contrast, the highly multimodal problems with medium or low global structure (F15-F24) use a comparatively much higher population size. As would be expected, for most functions, the used population size increases as dimensionality grows. For some functions, this trend is not maintained, which might indicate that the selected limit (200) was too low. This shows that the default population size in Table~\ref{tab:module-list} is indeed not ideal, and higher values should be considered instead. 

\begin{figure}[!t]
    \centering
    \includegraphics[width=0.49\textwidth,trim=10mm 60mm 10mm 75mm,clip]{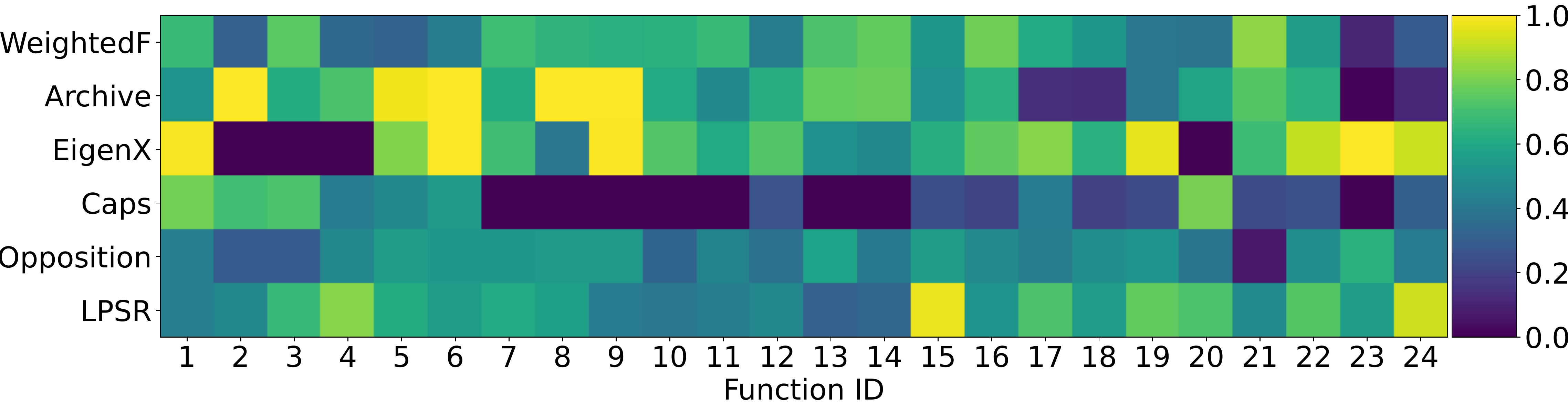}
    \caption{Fraction of elite configurations in which the specified binary module is on (aggregated across dimensions).}
    \label{fig:binary_modular_across_dim}
\end{figure}

\begin{figure}[!t]
    \centering
    \includegraphics[width=0.48\textwidth,trim=10mm 50mm 10mm 55mm,clip]{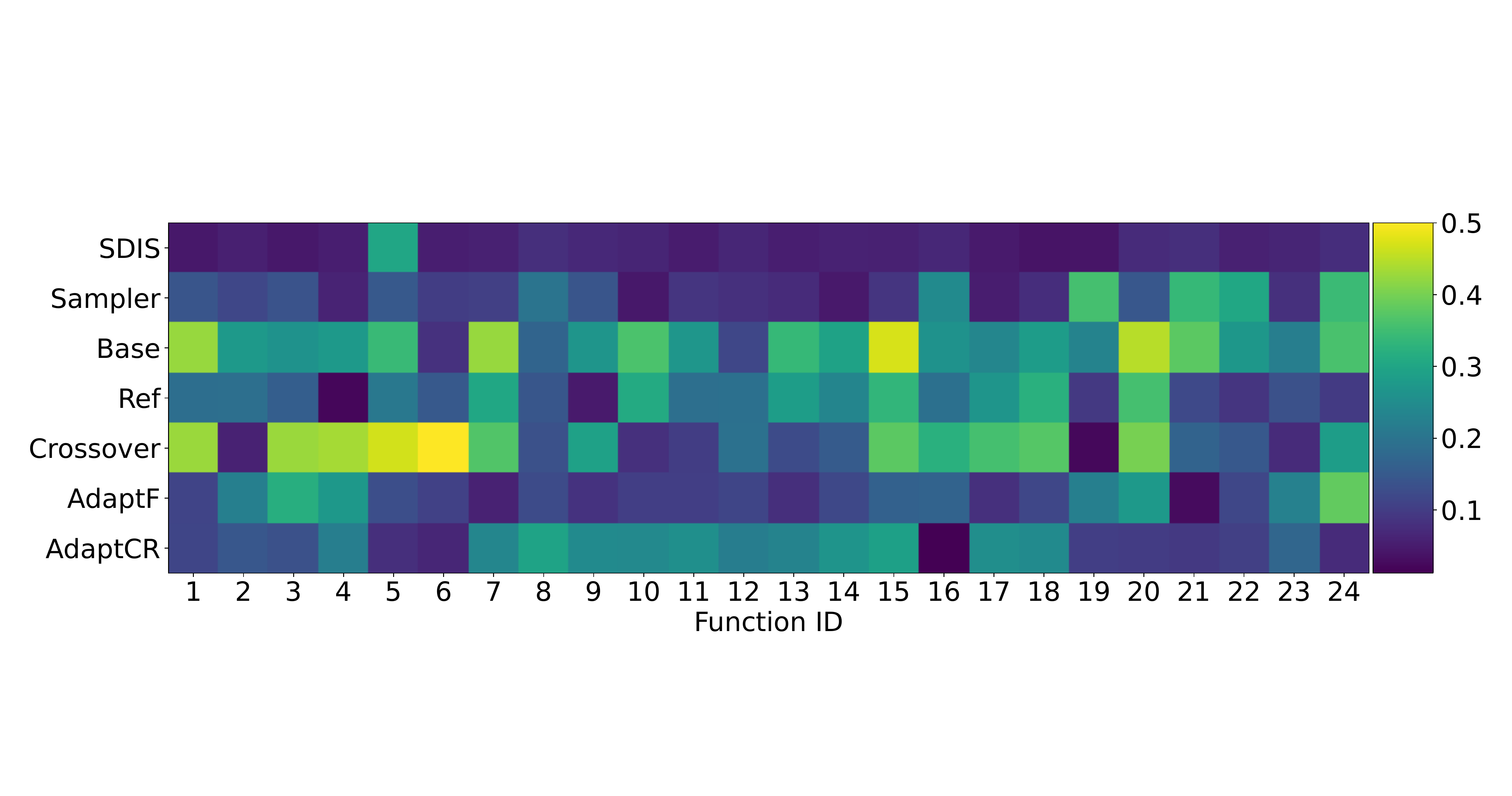}
    \caption{Standard deviation of module frequencies for selected modules.}
    \label{fig:module_std}
\end{figure}

Next to the parameter values, we can also consider patterns in the modules themselves. For the binary modules, this analysis results in Figure~\ref{fig:binary_modular_across_dim}, where we show the fraction of times a given module was activated in the elite configuration for each BBOB function. While values close to $0.5$ do not provide much direct insight, the more extreme values are interesting to consider. In particular, we notice that the capping mechanism for mutation as used in JSO (\texttt{Caps}) is rarely enabled, indicating that it appears to be detrimental to performance in most problems. 

\begin{figure*}
    \centering
    \includegraphics[width=0.99\textwidth,trim=12mm 21mm 5mm 11mm,clip]{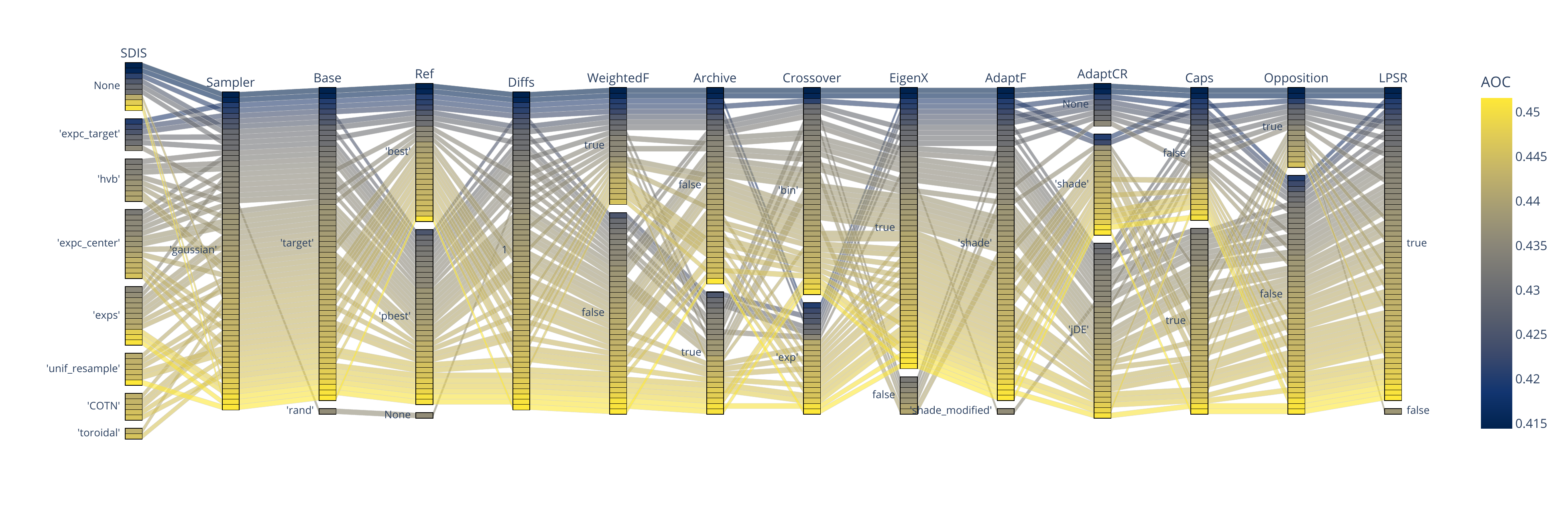}
    \caption{Parallel coordinate plot showing the modules activated in the elite configuration found across 10 runs of irace, on F19 in 5D. Configurations are colored based on normalized AOC.}
    \label{fig:example_parallel}
\end{figure*}

\begin{figure*}
    \centering
    \includegraphics[width=0.98\textwidth,trim=12mm 21mm 5mm 11mm,clip]{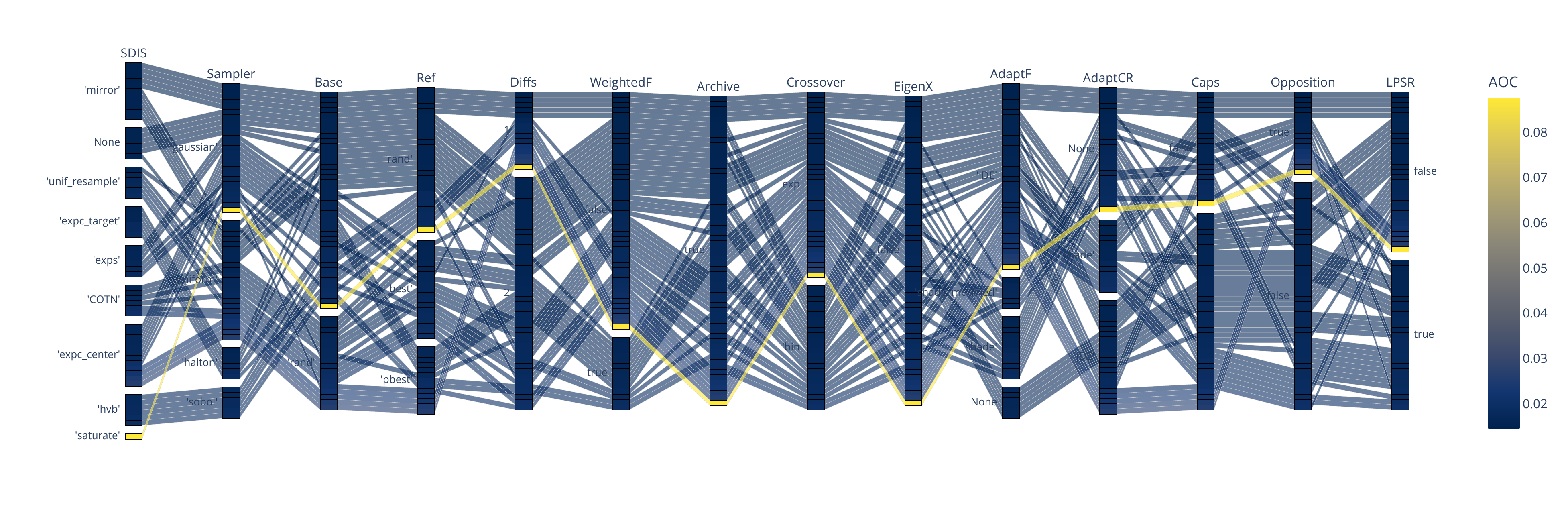}
    \caption{Parallel coordinate plot showing the modules activated in the elite configuration found across 10 runs of irace, on F2 in 10D. Configurations are coloured based on normalized AOC.}
    \label{fig:example_parallel2}
\end{figure*}

In addition to these binary modules, we can also analyse the other modules. This can be done by considering the standard deviation in the activation frequencies for each module. A low deviation corresponds to no changes from the mean, indicating that all modules are selected equally often, while high values (0.5 as maximum) indicate that one module is selected every time. These deviations, aggregated across dimensions, are visualised in Figure~\ref{fig:module_std}. From this figure, we see that the bound correction (\texttt{SDIS}) is distributed relatively uniformly, with the only exception being F5. This matches our previous observations in Figure~\ref{fig:single_module_importance_heatmap}, indicating that irace correctly identified the importance of this module during tuning. Furthermore, the mutation base (\texttt{Base}) seems to be a critical module, with large deviations from the initial uniform distribution. 

While Figure~\ref{fig:module_std} gives an overview of the distribution of modules, it does not show which options of these modules are selected more often. For this, we can instead consider the module activations of a single function and visualise them as a parallel coordinate plot. Figure~\ref{fig:example_parallel} shows this for Function 19 in 5D. In this figure, we see that all elite configurations make use of a Gaussian sampler for initialisation (\texttt{Sampler}). This makes sense when we consider the properties of F19 in more detail. In particular, we should note that for this function, the location of the optimum is not uniformly distributed in $[-4,4]^D$ as for most BBOB problems, but it is instead limited to the shell of the hypersphere of radius 1, centred at the origin~\cite{long2022bbob}. Because of this, a Gaussian initialisation will significantly outperform any uniform or low-discrepancy initialisation strategy. 

In Figure~\ref{fig:example_parallel} we also observe that all configurations, except one, make use of the SHADE-based adaptation for F (\texttt{AdaptF}), with `target' based mutation mechanism (\texttt{Base}). This suggests that, unlike the common belief of adding many components in the mutation operator to deal with such problems, adaptation systems based on the history of successful control parameter values are beneficial for multimodal problems similar to F19, especially when combined with `target'-based mutations and \texttt{Diffs}$=1$. 

We can compare these observations with those in Figure~\ref{fig:example_parallel2}, where we show the same visualisation for F2 in 10D. One significant difference between these two settings is the spread of performance. For F19, the differences between the best and worst configurations are greater than $0.1$, while for F2 these differences are at most $0.005$. This is partly explained by considering that F2 is a unimodal problem and, as such, the performance variability on this function is inherently lower than the one for multimodal F19. 
\section{Conclusions and Future Work}

In this paper, we introduced a modular framework for Differential Evolution and illustrate how it can help us gain insight into the differences between the many variations of DE proposed throughout the years. We show the flexibility of this framework by recreating a set of 11 \textit{common} versions of DE. Although these hand-crafted versions seem to outperform the default settings and other variations we create by changing only a single module, we see that the wide range of available options provides a wide potential to improve overall performance. 

 By utilising irace to tune the modules and parameters of this modular DE, we find consistent improvements over a set of \textit{common} DE variants. When adding more modules, we can repeat this process to gain an understanding of the newly created interaction with existing modules~\cite{de2021tuning}. We illustrated that the results from such a tuning process also help to gain further insight into the benefits of using different versions of DE for different problems. 

While the set of \textit{common} DE variants we used here by no means covers the full spectrum of hand-crafted DE versions that have been published, it shows another potential application of the proposed modular framework. If we can recreate an even larger set of previously benchmarked DE variants, we can take a large step towards a fair comparison of their individual contributions to the state-of-the-art. A large-scale benchmarking study using a modular framework would remove many aspects of inconsistency, resulting in a potentially more unbiased comparison. Such a study could show to what extent the published results in DE depend on factors such as the implementation and choice of the benchmark set-up. 

The modular DE we propose here is clearly not a complete framework that encapsulates all variations of DE. It might be a lofty goal to create one framework which covers this wide range of options, but striving towards coverage of more algorithm variants seems to be worthwhile. Although more modules increase the complexity of finding good configurations, the modular structure provides opportunities to use techniques such as knowledge graph embedding to create models that can effectively predict the performance of previously unused module combinations~\cite{kostovska2023using}. 

\bibliographystyle{ACM-Reference-Format}
\bibliography{references}

\end{document}